# A Novel Pixel-Averaging Technique for Extracting Training Data from a Single Image, Used in ML-Based Image Enlargement

**Amir Rastar**

*Auckland Bioengineering Institute, The University of Auckland, Auckland, New Zealand*
a.rastar@auckland.ac.nz
+64273086808
ORCID: 0000-0002-5271-529X

**Abstract**

Size of the training dataset is an important factor in the performance of a machine learning algorithms and tools used in medical image processing are not exceptions. Machine learning tools normally require a decent amount of training data before they could efficiently predict a target. For image processing and computer vision, the number of images determines the validity and reliability of the training set. Medical images in some cases, suffer from poor quality and inadequate quantity required for a suitable training set. The proposed algorithm in this research obviates the need for large or even small image datasets used in machine learning based image enlargement techniques by extracting the required data from a single image. The extracted data was then introduced to a decision tree regressor for upscaling greyscale medical images at different zoom levels. Results from the algorithm are relatively acceptable compared to third-party applications and promising for future research. This technique could be tailored to the requirements of other machine learning tools and the results may be improved by further tweaking of the tools' hyperparameters.

## 1. Introduction

In the evolution of artificial intelligence and machine learning, training data plays an important role in almost all of the performance measures used. Datasets used in different machine learning approaches vary from small to large sizes (e.g. big data), each of which needing different combination of tools to be dealt with. Amongst various attributes of a training dataset, size is always an important factor in determining the performance of the machine learning algorithm[1–3]. It is usually believed that, the bigger the size of the dataset, the more information it provides for the machine learning algorithm, leading to better outcomes. However, the availability of adequate data is subject to many different factors such as copyright protection, quality and accessibility of it. Not all the learning algorithms require large training data, however, it is important to provide just the right quantity to ensure the algorithm is efficiently functioning.

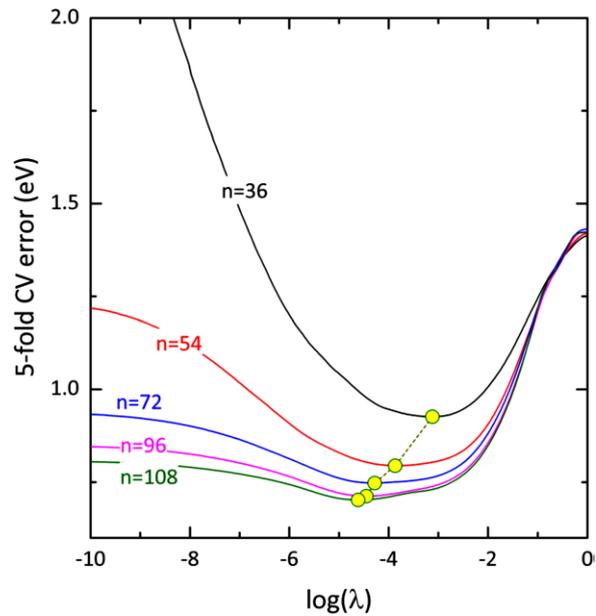
Figure 1. Effect of data size on cross validation RMSE of a LASSO regression. (λ) is the hyperparameter of the LASSO regression [3]

Image processing constitutes a substantial part of the problems addressed by machine learning experts. Many different tools are introduced in this field since the birth of machine learning and AI. Neural networks and different machine learning tools are more specifically used for image classifications. There are rapidly evolving thanks to powerful computing resources such as clouds and distributed computing clusters. Convolutional neural networks (CNNs) are an example AI tool widely used in image classification. For simpler purposes, binary and multi class classifiers could also be implemented to carry out the classification. All these tools require sufficient number of images to train the model prior to any prediction made by it [4–7]. I have developed an algorithm capable of creating the dataset for training using the properties of a single image which obviates the need for big datasets or even smaller ones which may not be accessible. I have called the algorithm pixel-averaging data generation algorithm which will be described in more detail in the following sections.

Throughout the history of medical imaging, engineers and researchers strived to introduce technological advancements in the field, whether for the imaging or post processing techniques, and they have succeeded in many ways. Recent implementation of AI and machine learning have augmented the imagining outcomes and for most of these tools, training dataset and its attributes again becomes a subject of interest [8–10].

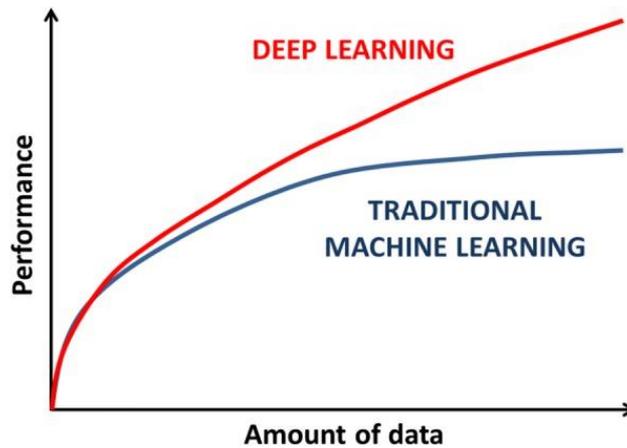
Figure 2. Comparison between classic ML and deep learning methods versus amount of data. Data size is a very important factor in model performance, particularly in newer Deep Learning tools [9].

Storing, archiving and transferring medical images are a crucial part of medical imaging techniques. Low-resolution images are easier to store and transfer, however, for practical purposes, they will need enhancements such as scaling, sharpening and histogram normalising before image registration and segmentation [11, 12]. Image processing techniques are widely used in medical image enhancement and recent developments of new machine learning and AI algorithms have integrated very well with the traditional techniques [13–18].

As of now, there hasn't been any literature about generation of sufficient data for training purposes of an intelligent medical image enhancement algorithm from a single image [19–22]. I have developed a technique with the ability of deriving enough training data from a single image to be used for relatively lossless upscaling of a medical image in a machine learning tool. Apparently, using multiple images with this technique will generate bigger training dataset, potentially leading to a better result from the algorithm, however, the purpose of this article is to present a fast and cost-effective technique as a good alternative when there is a lack of big or reliable training set. This tool could be used in different types of machine learning algorithms (Decision trees in this case) to train the algorithm for enlarging a poor-quality medical image at different zoom levels.

## 2. Materials and methods

Data used in this publication were generated by the National Cancer Institute Clinical Proteomic Tumor Analysis Consortium (CPTAC). Code used in the algorithm is in Python 3.6 developed in JetBrains PyCharm Community Edition. A simple decision tree was used without any regularisation of their hyperparameters for this study. The maximum number of leaves for the tree regressor was set to default. Scikit-learn library was used for implementing the regressors. After obtaining the prediction results from the regressors, a convolutional gaussian blur filter was applied twice to the image for noise removal.

## 3. Theory

An image is comprised of pixels. A slice of a medical image is usually a single channel greyscale image and the information is stored within a 0-255 scale range. I have presumed that the image dealt with, follows the same image format, hence, the pixel-averaging technique is applied on a grayscale single channel image. The idea of this technique is to divide the image is multiple sets of 4 pixels at multiple levels and calculating the average grey from each set. Assuming each set of 4 pixels yield an average, the final result of this averaging will be another greyscale image where its pixels are all average greys of 4-pixel sets of a bigger and higher resolution image. Proceeding to the next step, averaging is again applied to the generated image and this is continued recurrently until a single pixel is remained. If the initial image is not square, preliminary blocks of 16 pixels are extracted and averaged to 4 pixels as the first level 4-pixel set. This is performed for as many blocks as it is possible to generate from the original image. There might be a few pixel losses in this process, however as the pixels are only used for training purposes, the effect of loss should be negligible. Now starting from the final generated pixel and going back recursively, we can assume that each average pixel would generate a set of 4 pixels. Therefore, each average pixel could train a model to generate 4 pixels. This could be a good problem for multivariate regression and in this case a decision tree regressors was selected. Assuming that each pixel represents a set of 4 pixels, we have now constructed a dataset with one pixel as the feature and 4 pixels as the labels. Generating pixel averages from an image will provide enough training data for the desired ML algorithm.

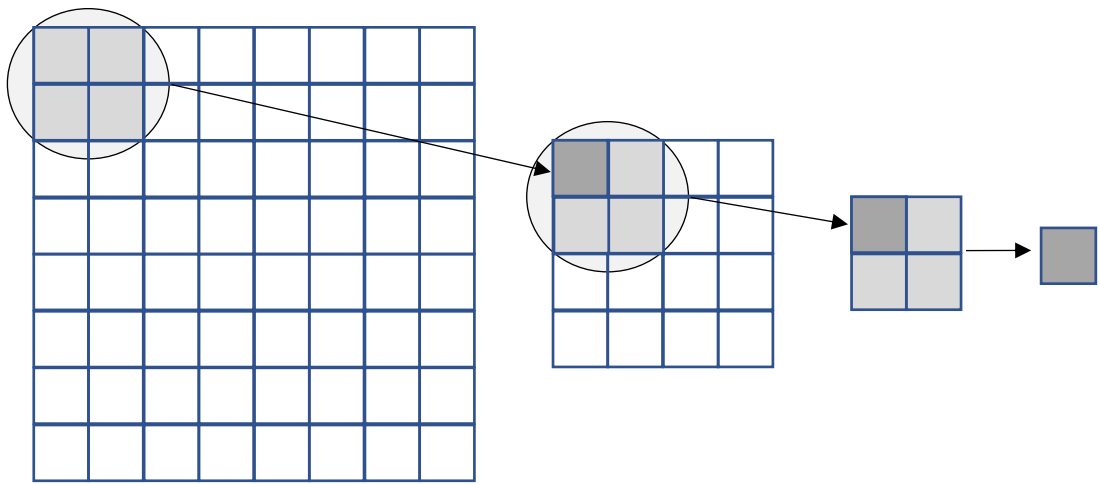

Figure 3. A schematic view of the pixel-averaging algorithm

If the pixels of the original image are now assumed to be the test set for the prediction, we can generate 4 pixels from each single pixel of the original image. This will scale up the original image 2 times (2X zoom) and this procedure could be applied to the generated images as many times as required by taking the generated image as the test set for prediction of the new upscaled image and so on.

## 4. Results

Results for 2, 3 and 4 times zooming of medical images are shown in figure 1 versus the original image. Similar upscaling levels were applied using third-party programs (Adobe Photoshop CC) to compare the performance of the algorithm with existing tools. Bicubic smoother enlargement method was used in the third-party app to carry out the scaling.

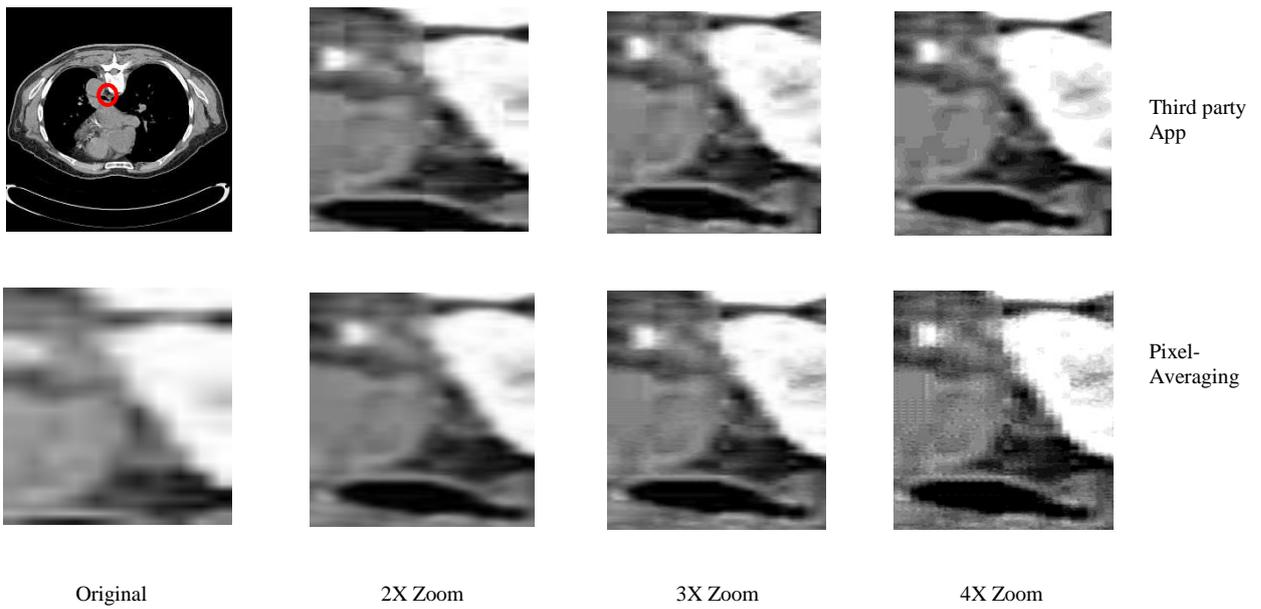

Figure 4. Comparison between different scaling levels of an image, (original image from National Cancer Institute Clinical Proteomic Tumor Analysis Consortium (CPTAC) [23])

Another comparison between two different scaling methods (pixel-averaging versus third-party scaling) could be observed in figure 2.

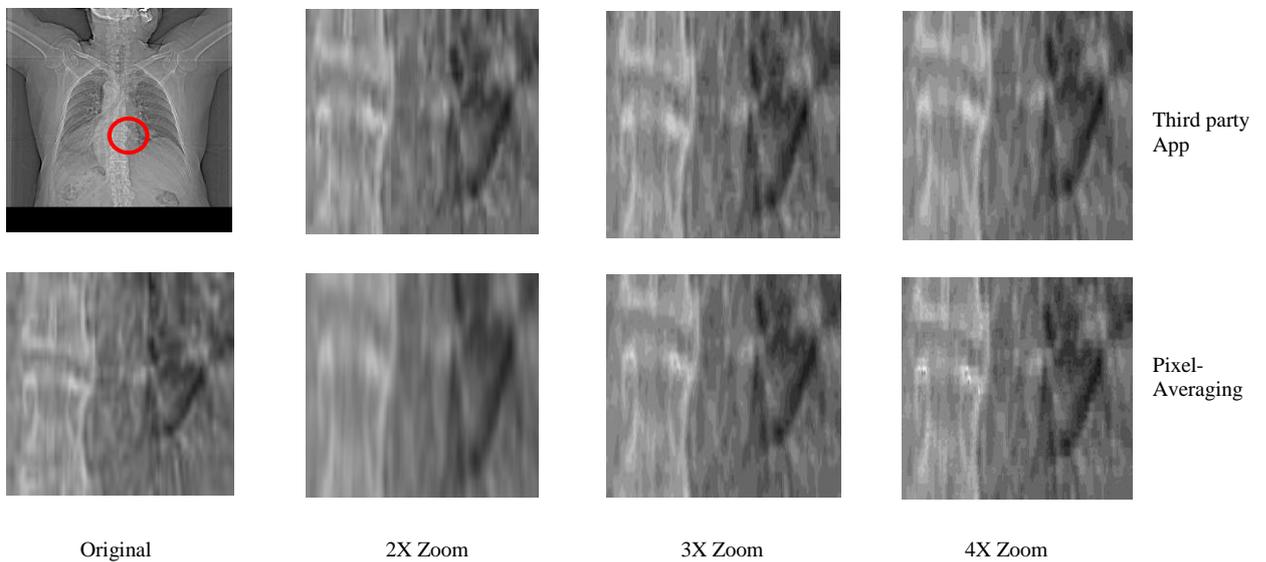

Figure 5. Comparison between different scaling levels of an image, (original image from National Cancer Institute Clinical Proteomic Tumor Analysis Consortium (CPTAC) [23])

Figure 4 and 5 show an acceptable image enlargement result from the pixel averaging ML algorithm compared to a commercial application, however, there is always room for more improvement.

## 5. Discussion

Results from the pixel-averaging algorithm is relatively acceptable in upsizing an image compared to the third-party application. The decision tree regressor wasn't regularized for their hyperparameters whereas, it could be optimised through simple tweaking tools such as KFold cross-validation and GridSearch CV. The accuracy of the algorithm was above 95% for all images therefore, further parameter tweaking may not be necessary as it possibly would not improve the accuracy scores.

Technically, predicting the upscaled images from the original image is a semi-unsupervised method as there is no target data to evaluate the accuracy of the final image and the prediction is solely based on pretraining with the data extracted from the same image. The accuracy score reported is only for the training set and the test set accuracy could not be evaluated.

It is important to mention that an already high-resolution image would yield better results as more data could be extracted from it by nature. However, if there is no chance of finding a set of training images, there is no other option but adhering to this algorithm to generate data from the very single available image. Data augmentation methods for images such as scaling, rotating and etc could also be implemented on the same image as raw material for the pixel-averaging tool which is a possible way for adding information to the training dataset, especially for lower-resolution images.

## 6. Conclusion

I have introduced a novel pixel-averaging algorithm, capable of deriving enough information from a single greyscale medical image as the training set for machine learning algorithms used for image enlargement. The algorithm is at its very preliminary stage of development, nevertheless, it could generate acceptable results where a big training dataset of images is not accessible or reliable.

## Appendix

A sample code, original and modified pictures of this paper could be found at the address below:

https://github.com/arcisad/MLZOOM

## Acknowledgements

The Author would like to acknowledge Machine Learning Guru for providing the code for the noise removal convolution filter.

## Conflict of Interest

The authors declare that they have no conflict of interest.